\documentclass[letterpaper, 10 pt, journal, twoside]{IEEEtran}

\usepackage{amsmath}
\usepackage{amssymb}
\usepackage{bm}
\usepackage{graphicx}
\usepackage{subcaption}
\usepackage[skip=0pt,labelfont=bf]{caption}
\usepackage{wrapfig}
\usepackage{adjustbox}
\usepackage{pgfplots}
\usetikzlibrary{patterns}
\usepackage{color,soul}
\usepackage{amssymb}
\usepackage{url}
\usepackage{breqn}
\usepackage[top=57pt, left=48pt, right=48pt, bottom=43pt]{geometry}
\usepackage{cite} 

\begin{document}
%
\title{Modeling Grasp Type Improves\\ Learning-Based Grasp Planning}

\markboth{IEEE Robotics and Automation Letters. Preprint Version. ACCEPTED January, 2019}
{Lu \MakeLowercase{\textit{et al.}}: Modeling Grasp Type Improves Learning-Based Grasp Planning}  

%
%
%

\author{Qingkai Lu$^{1}$ and Tucker Hermans$^{1}$%
\thanks{Manuscript received: September 10, 2018; Revised December 5, 2018; Accepted January 3, 2019.}
\thanks{This paper was recommended for publication by Editor H. Ding upon evaluation of the Associate Editor and Reviewers' comments. 
Q.~Lu was supported by NSF Award \#1657596.}
\thanks{$^{1}$Qingkai Lu and Tucker Hermans are with the School of Computing and the Robotics Center, University of Utah, Salt Lake City, UT 84112, USA.
        {\tt\footnotesize qklu@cs.utah.edu; thermans@cs.utah.edu}}%
\thanks{Digital Object Identifier (DOI): see top of this page.}
}
\maketitle

\begin{abstract}
  Different manipulation tasks require different types of grasps. For example, holding a heavy tool like a hammer requires a multi-fingered power grasp offering stability, while holding a pen to write requires a multi-fingered precision grasp to impart dexterity on the object. In this paper, we propose a probabilistic grasp planner that explicitly models grasp type for planning high-quality precision and power grasps in real-time. We take a learning approach in order to plan grasps of different types for previously unseen objects when only partial visual information is available. Our work demonstrates the first supervised learning approach to grasp planning that can explicitly plan both power and precision grasps for a given object. Additionally, we compare our learned grasp model with a model that does not encode type and show that modeling grasp type improves the success rate of generated grasps. Furthermore we show the benefit of learning a prior over grasp configurations to improve grasp inference with a learned classifier.
\end{abstract}


\section{Introduction}
\label{sec:intro}
\IEEEPARstart{C}{ountless} robotic manipulation tasks require multi-fingered grasps due to the stability and dexterity they afford. It's necessary to plan different types of multi-fingered grasps to accomplish different kinds of manipulation tasks~\cite{cutkosky1989grasp, heinemann2015taxonomy}. For example, holding a heavy tool like a hammer or wrench requires a power grasp affording substantial stability, while holding a scalpel or pen requires a precision grasp providing increased dexterity. Most grasp learning work to date focuses on two-fingered grasping with parallel jaw grippers where grasp type is not a concern~\cite{gualtieri2015using, lenz2015deep,gualtieri2016high,pinto2016supersizing,levine2016learning,
mahler2017dex,johns2016deep,varley2015generating}. Existing multi-fingered grasp planning work tends to only produce grasps of a single type for a given object~\cite{varley2015generating, veres2017modeling, lu2017grasp, miller2003automatic, roa2012power, hang2016hierarchical, zhu2004planning}. This is especially true for learning-based grasp planning methods~\cite{varley2015generating, veres2017modeling, lu2017grasp}. Two issues must be addressed in investigating multi-type, multi-fingered grasp learning: (1) the multi-finger grasp configuration has substantially higher dimension compared to two-fingered grasps; and (2) power and precision grasps are often very different for the same object making them difficult to predict by a single learner.

\begin{figure}[h]
  \centering
  \includegraphics [width=0.5\textwidth] {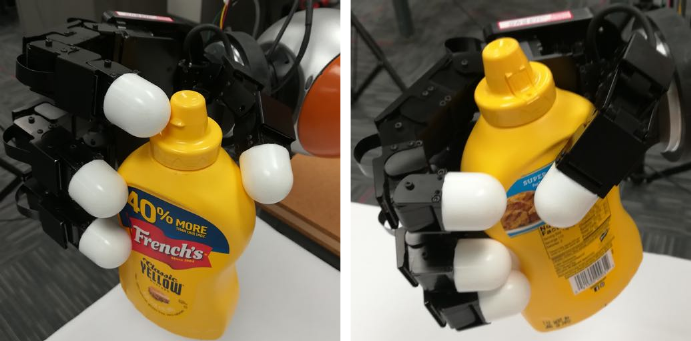}
  \caption{Precision (left) and power (right) grasps generated by our planner.}
  \label{fig:grasp_type_example}
\end{figure}
We overcome these challenges by introducing a novel probabilistic graphical model for grasp learning, which explicitly models grasp type. While our model can theoretically handle any number of grasp types, we focus on precision and power grasps in this letter.
Precision and power grasps are the two most common human and robotic grasp types~\cite{feix2016grasp,cutkosky1989grasp}. Following Cutkosky~\cite{cutkosky1989grasp} we define power grasps as having large areas of contact between the grasped object and the surfaces of the fingers and palm, and little or no ability to impart motions with the fingers. Precision grasps are defined by the hand holding the object with the tips of its fingers and thumb. In Figure~\ref{fig:grasp_type_example}, we show examples of precision and power grasps generated by our grasp planner for the same object.

Our probabilistic model for grasp type provides a unified framework for learning a function that predicts grasp success and for planning over grasp type and grasp configuration for a given object. We additionally introduce a data-driven prior over grasp configuration in order to improve inference for planning.
Our work is the first supervised grasp learning approach that can explicitly plan both power and precision grasps for a given object. We compare to a similar probabilistic grasp model that does not encode grasp type to show the benefit of explicitly modeling grasp type in planning grasps. Additionally, our experiments show that our model allows us to learn a high-quality predictor with far fewer data than a state-of-the-art deep neural network model~\cite{lu2017grasp}.

\section{Related Work}
\label{sec:related_work}
Robotic grasping approaches can generally be divided into analytical and learning-based methods~\cite{bohg2014data}~\cite{sahbani2012overview}. Analytical methods construct grasps that satisfy certain grasp properties, such as force closure, task compatibility, and other stability measures~\cite{sahbani2012overview}. Analytical approaches generate grasps based on geometric, kinematic, and/or dynamic object models~\cite{sahbani2012overview}. Learning-based methods use labeled grasping data to train classifiers to predict grasp success or regression to predict a desired grasp~\cite{bohg2014data, lu2017grasp}. Compared with analytical approaches, learning-based methods tend to generalize well to unknown objects with only partial object information~\cite{bohg2014data, lu2017grasp}.

Most analytical approaches~\cite{Grupen1991,ferrari1992planning} generate grasps that satisfy certain grasp properties without considering the grasp type. Both precision and power grasps can be force closure grasps. Analytical approaches in~\cite{miller2003automatic, roa2012power} plan power grasps and precision grasps in~\cite{hang2016hierarchical, zhu2004planning, Dai2015}.
Vahrenkamp et al.~\cite{vahrenkamp2018planning} propose a grasp planner to plan both power and precision grasps that generates grasp hypotheses based on the global and local geometric surface features from the object mesh. The planner then evaluates if the candidate grasps ensure force closure. While it can generate both precision and power grasps in theory, all grasps shown in the paper are precision grasps. Morales and colleagues~\cite{morales2006integrated} present a shape primitive based grasp planner similar to~\cite{miller2003automatic} to plan both precision and power grasps in a kitchen environment. A set of precision and power grasp strategies are defined for each shape primitive in the shape library. Different control strategies enable the planner to move to the desired preshape for power and precision grasps.

Learning-based grasp planning has become popular over the past decade~\cite{Saxena-aaai2008, kappler2015leveraging, gualtieri2015using}.
Recent grasp learning approaches focus on using deep neural networks to predict grasps for two fingered grasps of parallel jaw grippers in~\cite{lenz2015deep,gualtieri2016high,pinto2016supersizing,levine2016learning, mahler2017dex,johns2016deep,redmon2015real,kumra2016robotic}, where grasp type is not a real concern. Relatively fewer deep-learning approaches examine multi-fingered grasping~\cite{varley2015generating, veres2017modeling, lu2017grasp}. Unlike our approach, these existing learning-based, multi-fingered grasp planners cannot explicitly plan precision or power grasps, generating only power or precision grasps for given objects. In~\cite{osa2016experiments} Osa et al.\ use contextual reinforcement learning to learn grasping policies to generate different grasp types. They transfer the learned grasping motions to novel objects by selecting local surface features on the novel object similar to those seen during learning. While this method learns multiple grasp types, the types are selected based on the geometry of the grasp. As such it cannot explicitly enforce a specific grasp type be selected, a characteristic of our system.

In addition to object attributes, task requirements dictate grasp choice for a given hand~\cite{cutkosky1989grasp, cai2016understanding}. As such, some existing grasp planners incorporate constraints and preferences according to different manipulation tasks~\cite{dang2014semantic, song2015task, Fang2018Task}.  These methods plan grasps for different tasks, but they do not explicitly plan grasps of desired grasp types. These methods could potentially be combined with our approach to select the grasp type most appropriate for the desired task either through learning, planning, or by being directly selected by a human user.
Song et al. propose an imitation learning approach for robots to learn grasps for different tasks from human demonstrated grasps~\cite{song2015task}. However, the object representation used are hand-coded object properties (e.g. object class and size) and not visual features, meaning the grasp planner can only work with a limited set of known objects, unlike our work which can plan grasps for novel objects.
In~\cite{Fang2018Task}, a Task-Oriented Grasping Network (TOG-Net) is proposed to optimize a reaching and grasping policy for tool use. Similar to our approach, separate TOG-Nets are trained for each of the two tool tasks: sweeping and hammering.

\section{Modeling Grasp Types for Learning and Inference}
\label{sec:modeling_learning}
\begin{figure}[h]
 \vspace{-8pt}
    \centering
    \begin{subfigure}[t]{0.23\textwidth}
      \centering
        \includegraphics[width=\textwidth]{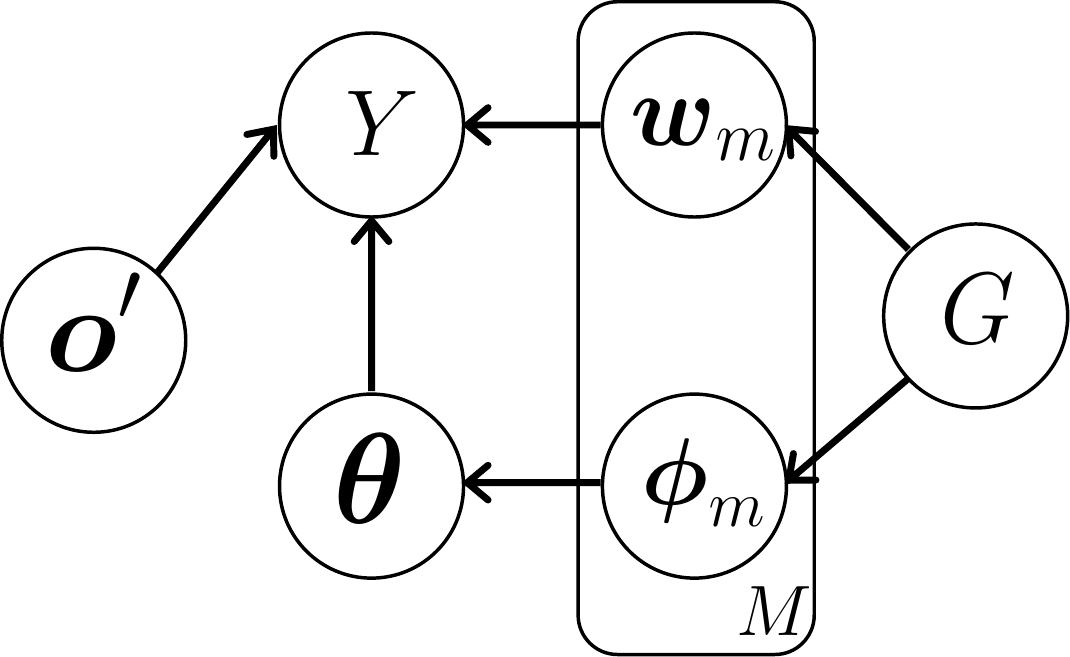}
        \caption{Our proposed grasp-type probabilistic graphical model.}
        \label{fig:pgm_with_type}
    \end{subfigure}
    ~ 
    \begin{subfigure}[t]{0.23\textwidth}
      \centering
        \includegraphics[width=\textwidth]{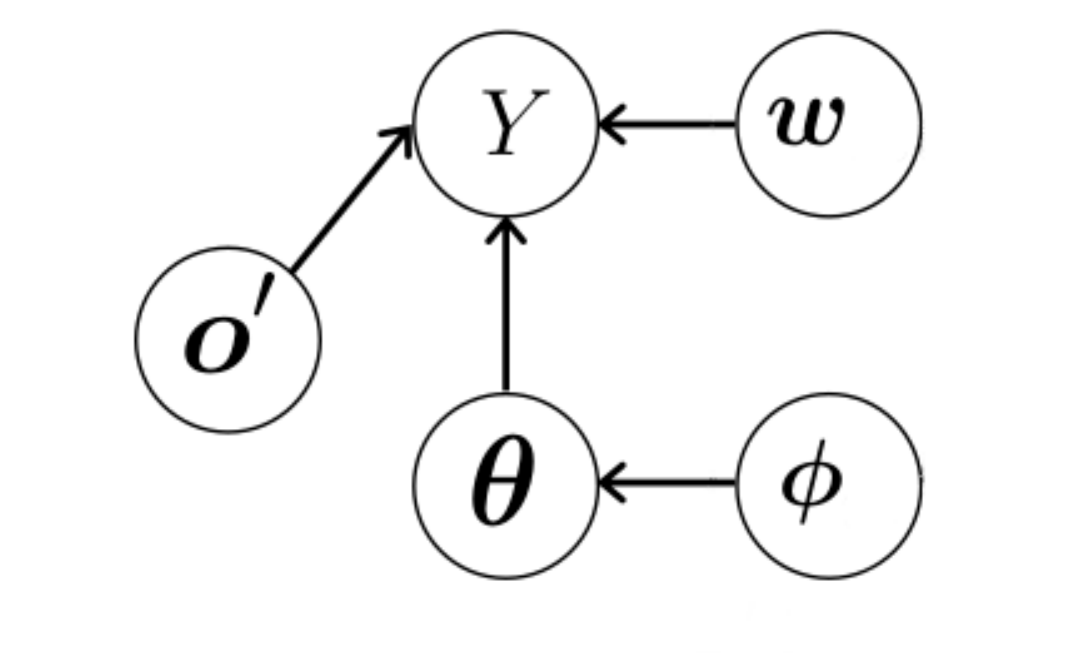}
        \caption{The ``type-free'' grasp model lacking a grasp type variable.}
        \label{fig:pgm_without_type}
    \end{subfigure}
    \caption{Bayes net of grasp probabilistic models with and without grasp types. $M$ is the number of grasp types.}
    \label{fig:grasp_pgm}
    \vspace{-5pt}
\end{figure}

In order to combine learning a classifier to predict grasp success with grasp synthesis in a principled manner, we propose a probabilistic graphical model which incorporates grasp type. By building such a model we can perform grasp planning as probabilistic inference as previously proposed~\cite{lu2017grasp,Saxena-aaai2008}.
We present our probabilistic graphical model for explicitly modeling grasp types in Fig.~\ref{fig:pgm_with_type}. This graphical model represents a Bayesian network---a directed, acyclic graph structure where each node represents a random variable and edges denote conditional independence in the joint probability distribution of all variables in the graph\cite{KollerPGMs2009}.

We have designed the structure of our graphical model to encode the generative, causal structure of the grasping process. At a high level our model encodes that grasp success depends directly on the specific object being grasped and the selected configuration of hand wrist and joint positions. Since grasp types define qualitatively distinct ways of performing grasps, we propose learning type-specific classifiers for predicting grasp success.

Formally, \(G\) defines the discrete grasp type variable using a categorical distribution, \(p(G)\), \(Y\) defines the binary random variable associated with grasp success or failure, given the grasp configuration parameters, \(\boldsymbol{\theta}\), the observed visual features \(\boldsymbol{o'}\), and the type-specific classifier weights \(\bm w_m\). This has the associated conditional probability distribution~$p(Y | \bm{\theta}, \bm{o'}, \bm{w}_m)$. We additionally introduce a data-driven prior, \(\bm{\phi}_m\), over grasp configuration for each grasp type in order to encode preferred grasp configurations independent of the observed object. $M$ defines the number of grasp types, and the plate under the classifier and prior parameters encodes that we have different values of those variables for each grasp type.

Fig.~\ref{fig:pgm_without_type} defines our comparison grasp model, that does not model grasp type, but does include our novel grasp configuration prior. We refer to this model without grasp types as the ``type-free'' model. This is equivalent to setting \(M=1\) in our model.

At training time the values of \(G\), \(\boldsymbol{o'}\), \(\boldsymbol{\theta}\), and \(Y\) are observed and the classifier weights \(\bm w_m\) and grasp prior parameters, \(\bm{\phi}_m\), are learned. Given these learned parameters we can predict the probability of grasp success, \(p(Y)\) for a given object using the observed visual features, \(\boldsymbol{o'}\), and specified grasp parameters, \(\boldsymbol{\theta}\), for each grasp type, \(m \in \{1,\ldots,M\}\).
However, when performing grasp planning we only observe the visual features, \(\boldsymbol{o'}\), and must infer the grasp configuration, \(\boldsymbol{\theta}^{*}\), that maximizes the probability of grasp success, \(p(Y)\).

\begin{figure*}[ht!]
  \centering
  \includegraphics [width=0.95\textwidth] {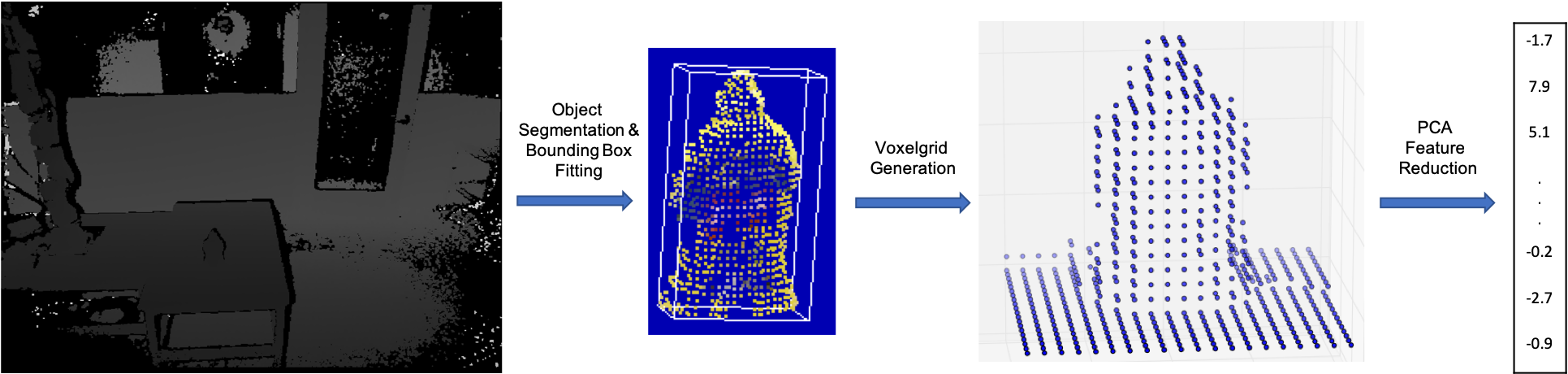}
  \caption{Illustration of the visual feature extraction process.}
  \label{fig:feature_diagram}
  \vspace{-15pt}
\end{figure*}


We use logistic regression to define the grasp success classifier in our model. The parameters $\bm{w}_m$ define the weights and bias for the classifier associated with grasp type $m$. A deterministic distribution defines the mapping from grasp type to the associated model parameters: \(p(\bm{w}_m|g) = 1\) if \(g = m\) and 0 otherwise. The same distribution is used for \(p(\bm\phi_m|g_m)\). Note one can use other models for the grasp classifier, such as a neural network as done in~\cite{lu2017grasp}. However, we found this model allowed us to learn with fewer data when using the visual features defined in Section~\ref{subsec:vis-feats}. We use an uninformative prior for \(p(G)\), asserting that all grasp types have equal prior probability.

We use Gaussian mixture models (GMM) to model the grasp configuration prior $p(\bm{\theta} | \bm \phi_m)$ for each grasp type \(m\). As such, \(\phi_m = \left\{\pi_{mk}, \bm{\mu}_{mk}, \bm{\Sigma}_{mk} \right\}\) is comprised of respectively the mixing weights, mean, and covariance of the \(k\) Gaussian components of the GMM associated with grasp type $m$. We introduce this grasp prior to constrain the inference used at planning time to not stray into areas far from grasp configurations observed at training time, where we have little evidence to support grasp success predictions.
\vspace{-12pt}
\subsection{Grasp Learning}
\label{subsec:grasp_learning}
Given a training set of observed samples \(T = \{(\bm\theta_i, g_i, \bm{o'}_i, y_i); i=1,\ldots,N\}\), we can use maximum likelihood estimation (MLE) to learn the grasp model parameters \(\bm{W}\) and \(\bm{\Phi}\). $\bm{W}$ represents the vector containing all grasp classifier weights and $\bm{\Phi}$ defines the vector for all grasp prior parameters.
The MLE parameter estimation objective function is formulated as:
\begin{align}
& \bm{W}^{*}, \bm{\Phi}^{*} \notag \\
&=\underset{\bm{W}, \bm\Phi}{\text{argmax}}
 \log  \prod_{i=1}^{N} p(\boldsymbol{\theta}_i, g_i, y_i, \boldsymbol{o'}_i, \bm w, \bm \phi) \label{eq:learn_obj}\\
&= \underset{\bm W, \bm\Phi}{\text{argmax}}
 \log \prod_{i=1}^{N}  p(y_i | \boldsymbol{\theta}_i, \boldsymbol{o'}_i, \boldsymbol{w}_m) \notag \\
& \hspace{75pt} p(\boldsymbol{w}_m|g_i)p(\boldsymbol{\theta}_i |\bm\phi_m)p(\bm\phi_m|g_i) \label{eq:bayes-net-model}\\
&= \underset{\bm W, \bm\Phi}{\text{argmax}}
 \sum_{i=1}^{N} \left[ \log p(y_i | \boldsymbol{\theta}_i, \boldsymbol{o'}_i, \boldsymbol{w}_m) + \log p(\boldsymbol{\theta}_i | \bm\phi_m) \right]_{m=g_i} \label{eq:log-prob-learning}
\end{align}
Where Eq.~\ref{eq:bayes-net-model} applies the factorization to the joint probability defined by our grasp-type graphical model.
The subscript \(m=g_i\) in Eq.~\ref{eq:log-prob-learning} denotes that variable \(m\) takes the value of the grasp type observed for sample \(i\). Combining this with the fact that, for a given sample \(i\), the two terms in the summand pertain separately to the classifier and grasp prior parameters, gives us the following learning criteria:
\begin{align}
  \bm{w}_m^{*} &= \underset{\bm w_m}{\text{argmax}}
               \sum_{i=1}^{N} \log p(y_i | \boldsymbol{\theta}_i, \boldsymbol{o'}_i, \boldsymbol{w}_m) &\forall g_i = m\label{eq:opt_classifier}\\
  \bm{\phi}_m^{*} &= \underset{\bm{\phi}_m}{\text{argmax}}
               \sum_{i=1}^{N} \log p(\boldsymbol{\theta}_i | \bm\phi_m) &\forall g_i = m \label{eq:opt_prior}
\end{align}
enabling us to independently learn the parameters for each grasp-type-specific classifier and prior.

We use coordinate descent to learn the parameters of our logistic regression classifiers. We fit the GMM parameters using the EM algorithm. The GMM prior has $4$ Gaussian components for both the type-based and type-free models in this letter. The grasp configuration vector, \(\bm\theta\), is composed of the palm pose in the object frame and the hand's preshape joint angles that define the shape of the hand prior to closing to execute the grasp. We explain the specific joints used for defining the preshape for our experiments at the beginning of Section~\ref{sec:data_collection}.

\subsection{Visual Feature Extraction}
\label{subsec:vis-feats}

We use a PCA-reduced voxel grid representation to encode the object visual features \(\bm o'\).
We illustrate the steps of our object visual features extraction in Figure~\ref{fig:feature_diagram}.
We first segment the object from the 3D point cloud by fitting a plane to the table using RANSAC and extracting the points above the table.
  We then estimate the first and second principle axes of the segmented object to create a right-handed reference frame, visualized as a bounding box in the second image from the left in Figure~\ref{fig:feature_diagram}. We then generate a $20\times20\times20$ voxel grid oriented about this reference frame. We define the center of the voxel grid to be the centroid of the points in the object segmentation.
Each voxel has size $0.01m\times0.01m\times0.01m$. A voxel grid with these dimensions covers the training and test objects we used and typically includes part of the table top. We find it helps the grasp planner to have information about the table top.

We then apply PCA to reduce the number of voxels from $8000$ to $15$ dimensions for each object. This 15 dimension vector serves as the visual features \(\bm o'\).
Our voxel encoding is inspired by~\cite{burchfiel2017eigen}, who constructs an object representation used for object classification, pose estimation, and 3D geometric completion using Variational Bayesian Principal Component Analysis (VBPCA). We first examined using VBPCA instead of PCA to encode the object shape, as in~\cite{burchfiel2017eigen}, but found it to have similar results at a much greater computational cost.

We estimate the PCA projection matrix using both successful and failed grasps of all grasp types used in our training set. We performed cross-validation to determine the PCA dimension looking at the prediction accuracy and F1 score. We evaluated an increasing number of latent dimensions from $5$ to $8000$ and find that performance does not significantly change after having $15$ dimensions. As such we keep $15$ for all experiments in this letter.


\vspace{-12pt}
\subsection{Probabilistic Grasp Inference}
\label{sec:inference}
Given the learned model parameters, \(\bm W\) and \(\bm\Phi\), we can perform grasp planning as probabilistic inference in our graphical model.
Given the features \(\bm{o'}\) associated with an observed object of interest, our goal is to infer both the grasp type, \(G\), and the grasp configuration parameters, \(\bm\theta\), that maximize the probability of success \(Y=1\). We formally define this inference as:
\begin{flalign}
\underset{\boldsymbol{\theta}, g}{\text{argmin}}\hspace{15pt}&-\log p(\boldsymbol{\theta}, g | y=1, \boldsymbol{o'}; \bm{W}, \bm\Phi) \\
\text{subject to}\hspace{8pt}&{\bm\theta}_{min} \leq \bm\theta \leq {\bm\theta}_{max}
\label{eq:inf_obj}
\end{flalign}
Where we constrain the grasp configuration parameters to obey the joint limits of the robot hand in Eq.~\ref{eq:inf_obj}.
Our formulation is similar to that used in~\cite{lu2017grasp}; however, our approach only has constraints on the hand joint limits, while~\cite{lu2017grasp} restricted the hand to stay within a local bounding box of the initial wrist pose in addition to the constraints of hand joint limits.
We can relax this constraint as our learned grasp configuration prior, keeps the inference from moving into unexplored regions of grasp configuration space. We assume each grasp has an equal prior probability of being power or precision.
 While we only examine two grasp types in this letter our inference algorithm naturally handles multiple grasp types and could easily be adapted to handle non-uniform priors over grasp type. We derive our objective function for inference starting in Eq.~\ref{eq:inf_obj_derive}.
\begin{align}
&\bm\theta^*, g* \notag \\
&= \underset{\boldsymbol{\theta}, g}{\text{argmin}} -\log p(\boldsymbol{\theta}, g | y=1, \boldsymbol{o'}, w, \bm\phi) \label{eq:inf_obj_derive}\\
& = \underset{\boldsymbol{\theta}, g}{\text{argmin}} - \log \prod_{m=1}^{M}  \Big[ p(y=1 | \boldsymbol{\theta}, \boldsymbol{o'}, \boldsymbol{w}_m)p(\boldsymbol{w}_m|g) \notag \\
&\hspace{90pt} p(\boldsymbol{\theta} | \bm\phi_m)p(\bm\phi_m|g) \Big] \\
& = \underset{\boldsymbol{g}}{\text{argmin}} \Bigl\{\underset{\boldsymbol{\theta}}{\text{argmin}}  - \log \Big[p(y=1 | \boldsymbol{\theta}, \boldsymbol{o'}, \boldsymbol{w}_{m=g}) \Bigr. \notag\\
& \hspace{90pt} \Bigl.  p(\boldsymbol{\theta} | \bm\phi_{m=g}) \Big] \Bigr\}\\
& = \underset{\boldsymbol{g}}{\text{argmin}} \Bigl\{\underset{\boldsymbol{\theta}}{\text{argmin}} - \log \Big[ p(y=1 | \boldsymbol{\theta}, \boldsymbol{o'}, \boldsymbol{w}_{m=g}) \Bigr. \notag \\
& \hspace{90pt} \Bigl. - \log p(\boldsymbol{\theta} | \bm\phi_{m=g}) \Big] \Bigr \}
\end{align}

In order to solve this nested pair of optimizations, we initialize and solve the inner optimization over grasp configuration \(\bm\theta\), for each grasp type \(g\). We then can select the grasp configuration \(\bm\theta^*\) and grasp type \(g^*\) that attained the minimum value in the inner optimization. Applying our model of a logistic regression classifier and GMM prior we obtain the following objective for this inner, grasp-type specific optimization:
\begin{align}
f(\bm\theta,g) = - \log \left(\frac{1}{1 + \exp(-\boldsymbol{w}_{m=g}^{T}\boldsymbol{x})}\right) - \notag \\
 \log \sum_{k=1}^{K}\pi_{k}\mathcal{N}(\boldsymbol{\theta}|\boldsymbol{\mu}_{m=g,k}, \boldsymbol{\Sigma}_{m=g,k}) \label{eq:grasp-inf-log-gmm}
\end{align}
Where \(\bm x\) defines the full set of features input to the classifier (i.e. the concatenated visual features, \(\bm{o'}\) and grasp configuration, \(\bm\theta\)).
The associated gradient for inference is:
\begin{align}
 \frac{\partial f(\bm{\theta}, g)}{\partial \bm\theta} = -\boldsymbol{w}_{g\theta} [ 1 - \frac{1}{1 + \exp(-\boldsymbol{w}_g^{T}\boldsymbol{x})}] +  \notag \\
\sum_{k=1}^{K} \frac{\pi_{k}
\mathcal{N}(\boldsymbol{\theta}|\boldsymbol{\mu}_{gk}, \boldsymbol{\Sigma}_{gk}) \mathbf{\Sigma}_{gk}^{-1}(\boldsymbol{\theta} - \boldsymbol{\mu}_{gk})}{\sum_{k=1}^{K}\pi_{k}\mathcal{N}(\boldsymbol{\theta}|\boldsymbol{\mu}_{gk}, \boldsymbol{\Sigma}_{gk})}
\label{eq:inf_obj_gradient}
\end{align}

In Eq.~\ref{eq:inf_obj_gradient}, $\boldsymbol{w}_{g\theta}$ denotes the classifier weights corresponding only to the grasp configuration parameters associated with grasp type \(g\) (i.e. we remove the sub-vector of weights associated with the visual features \(\bm{o'}\)). The inference objective function is non-convex with a GMM prior having more than one mixture components. As such, we can only hope to find a locally optimal grasp configuration. However, this is not a real issue, as many feasible grasps exist for commonly manipulated objects. We show in the next section, that given an initial grasp configuration close to the object, we tend to converge to a successful grasp.

We use L-BFGS with bound constraints~\cite{byrd1995limited}~\cite{zhu1997algorithm} to efficiently solve the MAP inference for each grasp type. We add a $0.5$ regularization parameter to the log prior term to prevent the prior dominating the inference. We use a heuristic grasp to initialize the inference, which we describe in detail in Section~\ref{sec:data_collection}. We use the scikit-learn (\url{http://scikit-learn.org/stable/index.html}) library to perform both learning and inference.


\begin{figure}[ht!]
  \vspace{-6pt}
  \centering
  \includegraphics [width=0.4\textwidth] {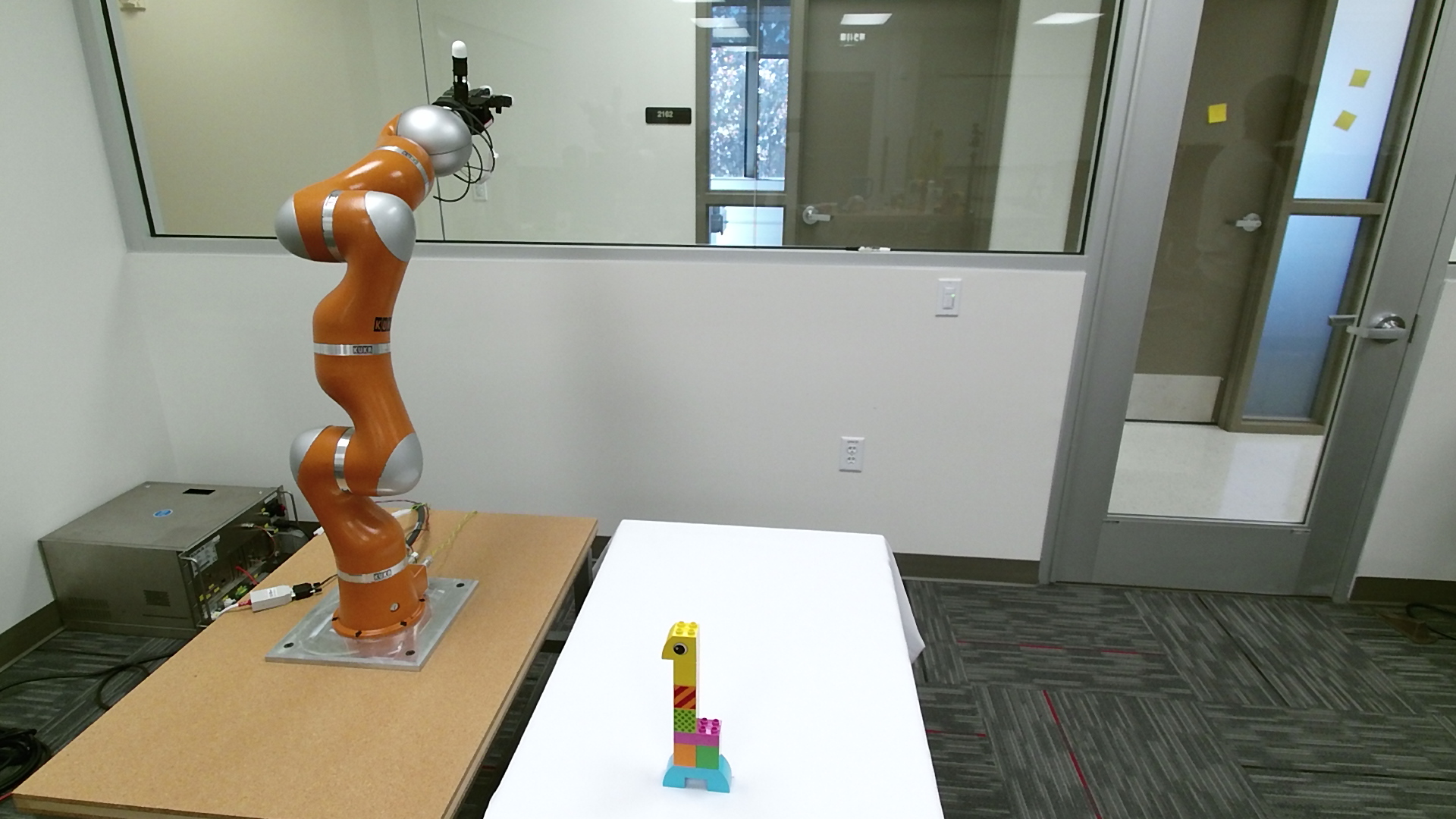}
  \caption{Example RGB image generated by the Kinect2 showing the robot and the ``lego'' object on the table.}
  \label{fig:kinect2_image}
  \vspace{-18pt}
\end{figure}
\section{Experimental Results}
\label{sec:experiments}
 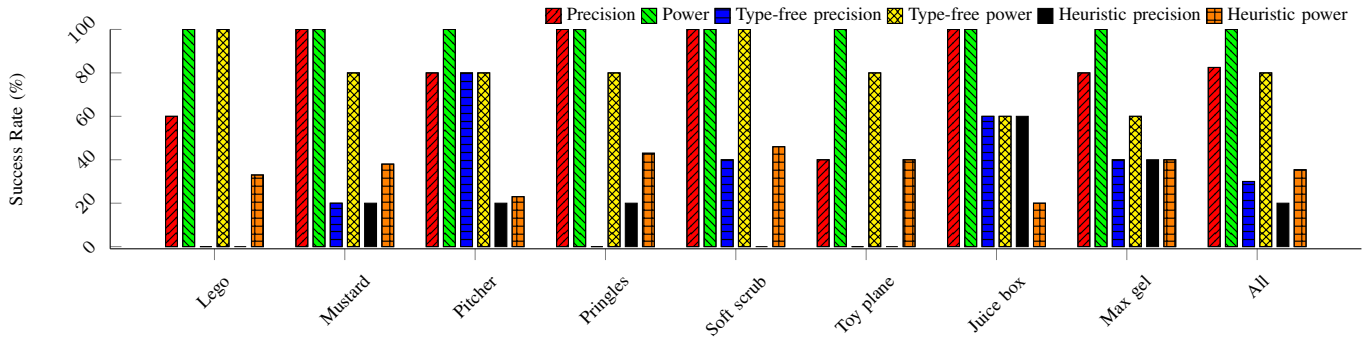
\begin{figure*}[ht!]
   \centering
  \begin{adjustbox}{minipage={\textwidth}}
    \begin{tikzpicture}
      \begin{axis}[
        ybar, ymin=-1, ymax=100,
        ylabel={Success Rate (\%)},
        axis lines*=left,
        symbolic x coords={Lego, Mustard, Pitcher, Pringles, Soft scrub, Toy plane, Juice box, Max gel, All},
        xtick=data,
        ticklabel style = {font=\scriptsize, rotate=45},
        legend style={font=\scriptsize, draw=none, fill=none},
        ylabel style = {font=\scriptsize},
        bar width = 4.5pt, height=4.5 cm, width=\linewidth,
        legend style={area legend, at={(1,1.15)}, anchor=north east, legend columns=6, },
        legend image code/.code={%
         \draw[#1] (0cm,-0.1cm) rectangle (2mm,1mm);},
        ]
        \addplot [fill=red, postaction={pattern=north east lines}] coordinates {
          (Lego, 60) (Mustard, 100) (Pitcher, 80) (Pringles, 100) (Soft scrub, 100)
          (Toy plane, 40) (Juice box, 100) (Max gel, 80) (All, 82.5)};

        \addplot [fill=green, postaction={pattern=north west lines}] coordinates {
          (Lego, 100) (Mustard, 100) (Pitcher, 100) (Pringles, 100) (Soft scrub, 100)
          (Toy plane, 100) (Juice box, 100) (Max gel, 100) (All, 100)};

        \addplot [fill=blue, postaction={pattern=horizontal lines}] coordinates {
          (Lego, 0) (Mustard, 20) (Pitcher, 80) (Pringles, 0) (Soft scrub, 40)
          (Toy plane, 0) (Juice box, 60) (Max gel, 40) (All, 30)};

        \addplot [fill=yellow, postaction={pattern=crosshatch}] coordinates {
          (Lego, 100) (Mustard, 80) (Pitcher, 80) (Pringles, 80) (Soft scrub, 100)
          (Toy plane, 80) (Juice box, 60) (Max gel, 60) (All, 80)};

        \addplot [fill=black] coordinates {
          (Lego, 0) (Mustard, 20) (Pitcher, 20) (Pringles, 20) (Soft scrub, 0)
          (Toy plane, 0) (Juice box, 60) (Max gel, 40) (All, 20)};

         \addplot [fill=orange, postaction={pattern=grid}] coordinates {
          (Lego, 33) (Mustard, 38) (Pitcher, 23) (Pringles, 43) (Soft scrub, 46)
          (Toy plane, 40) (Juice box, 20) (Max gel, 40) (All, 35.4)};

        \legend{Precision, Power, Type-free precision, Type-free power, Heuristic precision, Heuristic power}
      \end{axis}
    \end{tikzpicture}
  \end{adjustbox}
  \caption{Grasp success rates of different methods for multi-fingered grasping on the real robot. A grasp is counted as successful only if it leads to a successful lift of the object with the correct grasp type. All objects are previously unseen in training except the pringles can. All objects except the orange juice box and drano max gel are from YCB.}\label{fig:real_suc_rate}
  \vspace{-5pt}
 \end{figure*}
 

In this section, we describe our training data collection, grasp success classifier evaluation, and experimental evaluation of our grasp planner. We conduct experiments using the four-fingered, 16 DOF Allegro hand mounted on a Kuka LBR4 arm. We used a Kinect2 camera to generate the point cloud of the object on the table. An example image generated by the Kinect2 of the robot and an object can be seen in Figure~\ref{fig:kinect2_image}. We performed real-robot grasp experiments on $8$ objects spanning different shapes and textures. We compare our grasp planner with a type-free grasp planner to investigate the effects of explicitly modeling grasp type in learning. We additionally compare our grasp planner to a geometry-based heuristic grasp planner, which we also use for initialization of our grasp inference. In total the robot performed $240$ different grasps across all experiments. The code and data associated with this letter are available at: \url{https://robot-learning.cs.utah.edu/project/grasp_type}.

\subsection{Multi-finger Grasping Data Collection and Model Training}
\label{sec:data_collection}
We collected training data using the heuristic, geometry-based grasp planner from~\cite{lu2017grasp} which is quite similar to the geometric primitive planner of~\cite{miller2003automatic} for boxes or cylinders. Nearly all successful examples from~\cite{lu2017grasp} are power grasps. As such, we use the same objects (from the Bigbird data-set~\cite{singh2014bigbird}) and the same grasp data collection system used in~\cite{lu2017grasp} to collect additional multi-fingered precision grasps for our training data-set. For precision grasps we adapt the grasp generation protocol slightly from that in~\cite{lu2017grasp}, setting the palm pose to be a distance of $6cm$ from the center of the segmented object's bounding box face. We add noise to this pose by sampling from a Gaussian with a standard deviation of $2cm$. There are 14 parameters for the Allegro hand preshape, 6 for the palm pose and 8 relating to the first 2 joint angles of each finger proximal to the palm. Given a desired pose and preshape we use the RRT-connect motion planner in MoveIt! to plan a path for the arm. We execute all feasible plans moving the robot to the sampled preshape.

\begin{figure*}[ht!]
   \centering
   \includegraphics[ width=1.\linewidth]{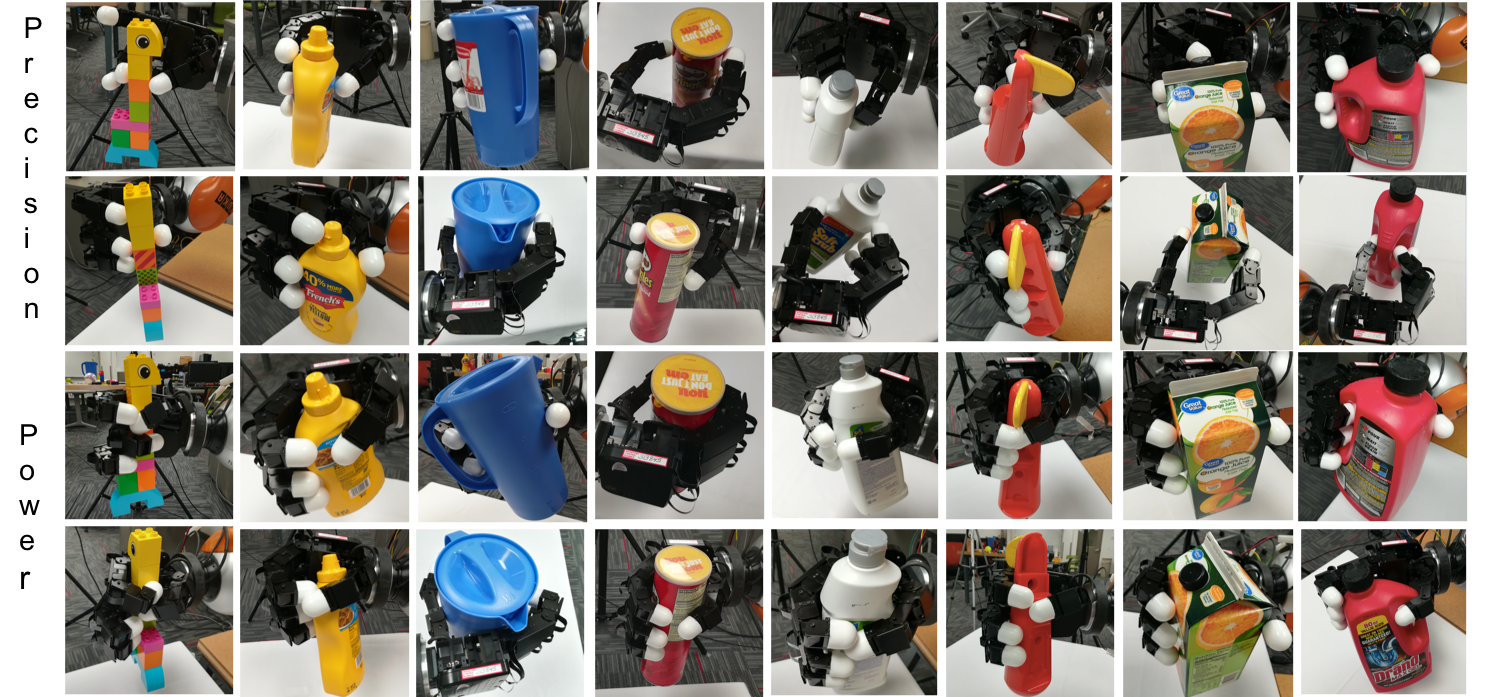}
   \caption{Examples of successful precision and power grasps generated by our grasp type modeling approach to grasp planning. Top two rows are precision grasps. Bottom two rows are power grasps.}
   \label{fig:grasp_examples}
   \vspace{-5pt}
 \end{figure*}
 After moving the hand to the desired preshape, a grasp controller is applied to close the hand. The grasp controller closes the fingers at a constant velocity stopping each finger independently when contact is detected by the measured joints velocities being close to zero. We use two different grasp controllers--one for executing precision grasps and one for performing power grasps. The power grasp controller attempts to envelop the object to generate stable power grasps with large contact areas between the object and the hand, while the precision grasp controller keeps the fingers more outstretched to generate dexterous precision grasps with fingertip contacts. We achieve this by having the power grasp controller close the three distal joints of the non-thumb fingers and the two distal joints of the thumb. The precision grasp controller closes the second and third joints of the non-thumb fingers and the two distal joints of the thumb. Note the proximal joint of all non-thumb fingers rotates the finger about its major axis causing it to change the direction of closing. As such we maintain the angle provided by the grasp planner for these joints.

If the robot grasps and lifts the object to a height of $15cm$ without the object falling, the simulator automatically labels the grasp as successful. We had to collect more than $1200$ precision grasps attempts in order to get our target number of $20$ successful precision grasps for training. This demonstrates the difficulty of the precision grasp-learning problem. Our final training set consists of $20$ successful examples for both precision and power grasps. Power grasps were randomly selected from the dataset we produced for our previous work~\cite{lu2017grasp}. The $20$ successful power grasps cover $20$ objects with different shapes, sizes, and materials, while the $20$ successful precision grasps cover $11$ different objects. We randomly draw $40$ failure grasps both of precision and power for a final training set of $120$ samples in total.

The precision, power, and type-free logistic regression classifiers are evaluated using leave-one-out cross validation. For the precision classifier leave-one-out accuracy is $0.85$ with an F1 score of $0.78$. The power classifier leave-one-out has an accuracy $0.87$ and a F1 score of $0.81$. The type-free classifier leave-one-out has an accuracy $0.82$ and a F1 score of $0.74$ for all grasps. The type-free classifier leave-one-out has an accuracy of $0.8$ for precision grasps and  an accuracy of $0.83$ for power grasps.
Overall, both the type-specific and type-free grasp classifiers perform reasonably well. However, the grasp classifiers for type-specific grasps outperforms the type-free grasp classifier in terms of both accuracy and F1 score.

\subsection{Real Robot Multi-finger Grasp Experiment}
We evaluated our grasp planner on the physical robot system. We performed experiments on 8 objects covering different geometries, sizes, and materials. We attempted grasps at 5 different poses per object, for a total of 40 grasp attempts per method. We use the same set of locations across different methods, but each object has its own set of random poses. We compare our grasp planner to the type-free grasp planner and a heuristic grasp planner. We evaluate both precision and power grasps for our approach and the comparison approaches. This results in a total of 6 different grasp trials (precision and power for the 3 different planners) giving a total of 240 grasp attempts on the physical robot.

We use the same heuristic precision grasp generation approach used in training data collection to generate heuristic grasps for testing. We use the heuristic grasp closest to the camera as the initial grasp configuration for grasp inference.  We apply the precision grasp controller to grasps generated with inferred precision grasp model parameters of our grasp-type planner and the power grasp controller for power grasp inference. We evaluate both precision and power grasp controllers for all grasps generated from the type-free grasp planner in order to give it the best chance of success.
Of the 8 objects used in experiments, 6 come from the YCB dataset~\cite{calli2015benchmarking}, 5 of which are the test objects used in~\cite{lu2017grasp}. Objects used in the experiments can be seen in Figure~\ref{fig:grasp_examples}.


We label a grasp attempt that successfully lifts the object to a height of $0.15m$ without dropping it a ``successful lift.''  A grasp is successful if (1) it is a successful lift and (2) its grasp type matches the planned grasp type. We define power grasps as those that have large areas of contact between the grasped object and the surfaces of the fingers and palm, and little or no ability to impart motions with the fingers; while precision grasps hold the object primarily with the tips of fingers and thumb~\cite{feix2016grasp}~\cite{cutkosky1989grasp}.
We treat grasps where the planner fails to generate a plan as failure cases. In our experiments all such cases would either collide with the object or were unreachable for the arm.

The grasp success rates for all methods are summarized in Figure~\ref{fig:real_suc_rate}. Our type-specific grasp planner has higher average success rates than the type-free grasp planner and the heuristic grasps for both precision and power grasps. For precision grasps, our grasp planner performs better than the baseline methods for all objects except ``pitcher'', where the type-free planner performs equally well. The only failed precision grasp for our method on the pitcher happens when it tries to grasp the handle for one object pose. We note that the learner saw no successful handle grasps in any training samples. The type-free planner was not able to plan precision grasps successfully for the ``lego'', ``pringles'', or ``toy plane'' objects. Our type-specific planner performs reasonably well in generating precision grasps on these three objects.

For power grasps, our grasp planner has a $100\%$ success rate for all objects tested. This perfect performance for power grasps from our grasp planner represents a higher success rate than the type-free grasp planner for $5$ of the $8$ objects tested and better than the geometric heuristic grasps for all objects. Our grasp planner successfully refined initial heuristics grasps which would fail into successful final power and precision grasps for all $8$ objects.


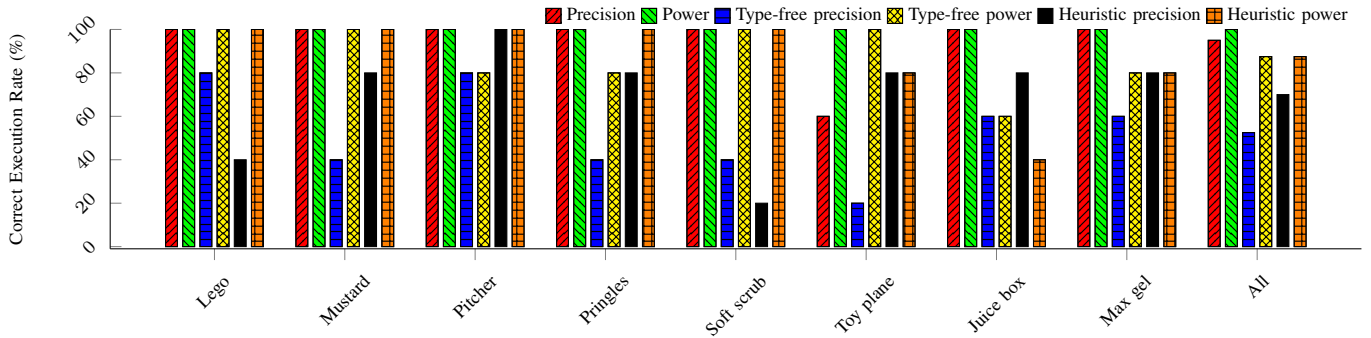
\begin{figure*}[ht!]
   \centering
  \begin{adjustbox}{minipage={\textwidth}}
    \begin{tikzpicture}
      \begin{axis}[
        ybar, ymin=-1, ymax=100,
        ylabel={Correct Execution Rate (\%)},
        axis lines*=left,
        symbolic x coords={Lego, Mustard, Pitcher, Pringles, Soft scrub, Toy plane, Juice box, Max gel, All},
        xtick=data,
        ticklabel style = {font=\scriptsize, rotate=45},
        legend style={font=\scriptsize, draw=none, fill=none},
        ylabel style = {font=\scriptsize},
        bar width = 4.5pt, height=4.5 cm, width=\linewidth,
        legend style={area legend, at={(1,1.15)}, anchor=north east, legend columns=6, },
        legend image code/.code={%
         \draw[#1] (0cm,-0.1cm) rectangle (2mm,1mm);},
        ]
        \addplot [fill=red, postaction={pattern=north east lines}] coordinates {
          (Lego, 100) (Mustard, 100) (Pitcher, 100) (Pringles, 100) (Soft scrub, 100)
          (Toy plane, 60) (Juice box, 100) (Max gel, 100) (All, 95)};

        \addplot [fill=green, postaction={pattern=north west lines}] coordinates {
          (Lego, 100) (Mustard, 100) (Pitcher, 100) (Pringles, 100) (Soft scrub, 100)
          (Toy plane, 100) (Juice box, 100) (Max gel, 100) (All, 100)};

        \addplot [fill=blue, postaction={pattern=horizontal lines}] coordinates {
          (Lego, 80) (Mustard, 40) (Pitcher, 80) (Pringles, 40) (Soft scrub, 40)
          (Toy plane, 20) (Juice box, 60) (Max gel, 60) (All, 52.5)};

        \addplot [fill=yellow, postaction={pattern=crosshatch}] coordinates {
          (Lego, 100) (Mustard, 100) (Pitcher, 80) (Pringles, 80) (Soft scrub, 100)
          (Toy plane, 100) (Juice box, 60) (Max gel, 80) (All, 87.5)};

        \addplot [fill=black] coordinates {
          (Lego, 40) (Mustard, 80) (Pitcher, 100) (Pringles, 80) (Soft scrub,20)
          (Toy plane, 80) (Juice box, 80) (Max gel, 80) (All, 70)};

         \addplot [fill=orange, postaction={pattern=grid}] coordinates {
          (Lego, 100) (Mustard, 100) (Pitcher, 100) (Pringles, 100) (Soft scrub, 100)
          (Toy plane, 80) (Juice box, 40) (Max gel, 80) (All, 87.5)};

        \legend{Precision,Power,Type-free precision,Type-free power,Heuristic precision,Heuristic power}
      \end{axis}
    \end{tikzpicture}
  \end{adjustbox}
  \caption{Correct execution rate of different grasp methods. Correct execution rate is the percentage of grasps that can be grasped and lifted up successfully with desired grasp types out of all grasps that can be grasped and lifted up successfully. All objects except the orange juice box and drano max gel are from YCB.}
  \label{fig:real_suc_wrong_type}
  \vspace{-15pt}
 \end{figure*}


Compared with the previous deep learning-based grasp inference of~\cite{lu2017grasp}, we show that our planner can plan both precision and power grasps with high quality while we require an order of magnitude fewer training examples. Moreover, the power grasps generated by our planner achieve a higher success rate than those in~\cite{lu2017grasp} for the same set of objects. 


Though we use separate precision and power grasp controllers, if the grasp planner fails to generate a good preshape for the desired grasp type, it can still execute a grasp with the wrong desired type (e.g. a desired precision grasp can become a power grasp and vice versa). In Figure~\ref{fig:real_suc_wrong_type}, we show the correct execution rate of the different grasp planners. Correct execution rate defines the percentage of successful lifts that achieve the desired grasp type across all successful lifts. As we can see from Figure~\ref{fig:real_suc_wrong_type}, our grasp planner correctly executes power grasps in all cases. Our grasp planner correctly executes all but two precision grasps correctly, executing the wrong grasp type twice on the ``toy plane'' (i.e. $95\%$ correct execution rate). The type-free planner correctly executes only $52.5\%$ of desired precision grasps that lift the object (i.e. $47.5\%$ successful lifts desired to be precision grasps become power grasps). The type-free planner achieves a much higher $87.5\%$ correct execution rate for power grasps (i.e. $12.5\%$ of successful lifts desired to be power grasps become precision grasps).
 For heuristic initialization grasps $70\%$ of successful lifts desired to be precision grasps have correct types.
 For heuristic initialization grasps $87.5\%$ of successful lifts desired to be power grasps have correct types.
 This demonstrates the preference for the type-free model to find power grasps compared to our type-specific grasp model.

While we generate some power grasps with the precision controller for the type-free and heuristic grasp planners, we cannot predict which type will result from the planned grasp for any given object.
As such, it makes it impossible to plan grasps of a specific desired type when necessary to satisfy task requirements. Moreover, we observe that power grasps generated using the precision grasp controller appear less stable and robust than power grasps generated by the power controller (e.g. the left grasp in Figure~\ref{fig:wrong_types_discussion}), and precision grasps generated from desired power grasps offer less mobility than those generated by the precision grasp controller (e.g. the right grasp in Figure~\ref{fig:wrong_types_discussion}).



In Figure~\ref{fig:grasp_examples}, we show two precision and power grasps for each object generated by our grasp planner. Grasps in the top two rows are precision grasps. Our precision grasps offer substantial mobility making them suitable for tasks like in-hand manipulation. Grasps in the bottom two rows show power grasps. These power grasps provide strong stability, making them good grasps for picking and placing objects. Qualitatively, we observe that grasps generated by our planner generally achieve higher quality (i.e. dexterity for precision grasps and stability for power grasps) than those generated by the two comparison methods.


\section{Discussion and Conclusion}
\label{sec:discussion}
In this letter, we present a probabilistic model for grasping that explicitly models grasp type. This enables us to learn type-specific grasp success classifiers, which in turn enable us  to plan high quality precision and power grasps as probabilistic inference. Our work is the first supervised grasp learning work that can explicitly plan both power and precision grasps for a given object. We compare to a similar planner that does not model grasp type and show through physical robot experiments that our grasp-type model achieves greater success in executing planned grasps.

\begin{figure}[ht!]
    \centering
    \begin{subfigure}[t]{0.23\textwidth}
      \centering
        \includegraphics[width=\textwidth]{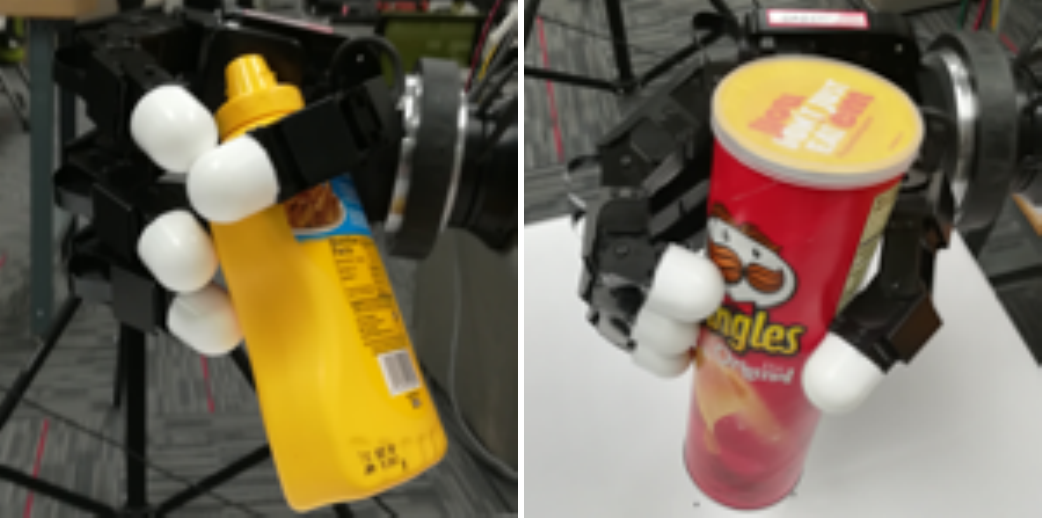}
        \caption{Resultant grasps differ from the intended grasp using the type-free planner.}
        \label{fig:wrong_types_discussion}
    \end{subfigure}
    ~ 
    \begin{subfigure}[t]{0.23\textwidth}
      \centering
        \includegraphics[width=\textwidth]{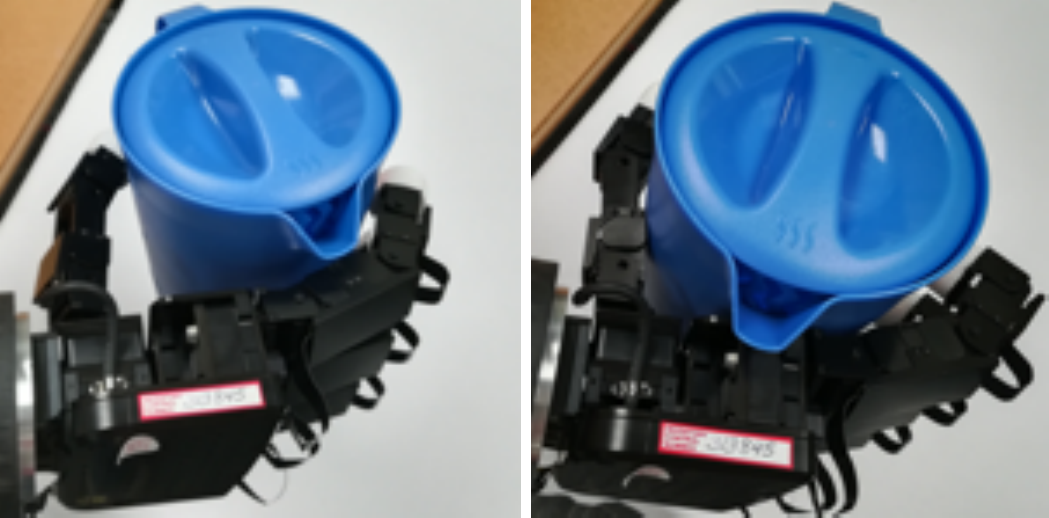}
        \caption{The type-free grasp model fails to predict resultant grasp type.}
        \label{fig:large_obj_discussion}
    \end{subfigure}
    \caption{Grasp examples for discussion.}
    \label{fig:results_discussion}
    \vspace{-15pt}
\end{figure}

The type-free grasp planner plans grasps with wrong types for most objects. We noticed the type-free grasp planner successfully plans and executes the correct grasp type for large objects with simple geometry, such as the ``pitcher'' and ``juice box.'' For example, in Figure~\ref{fig:large_obj_discussion}, the left image is a precision grasp of ``pitcher'' and the right image is a power grasp on the same object at the same pose. Since we can not predict the grasp type of the type-free planner before execution, it is impossible to intentionally plan grasps of a specific grasp type using the type-free planner as required for a given manipulation task.

In terms of maximum success probability our planner always preferred power grasps across all objects in our experiments. We attribute this to the power grasp prior probabilities for both the initialization and final inference being higher than for precision grasps. Since the same number of mixture components were used for both priors, this implies that the power grasps have lower overall variance (i.e. represent a smaller volume in grasp configuration space). We could potentially rectify this by providing more mixture components to the precision grasp prior.
Additionally, task requirements and object attributes together dictate the grasp choice~\cite{cutkosky1989grasp}. When humans grasp objects in everyday tasks, the grasp type choice is dictated less by the shape and size of the objects than the tasks they want to accomplish~\cite{cutkosky1989grasp}. As such, incorporating task-specific information could further inform grasp type selection.

As future work, we will collect a richer training set encompassing a larger diversity of both grasp types and object properties to evaluate our model with learned neural-network features. We additionally want to model more grasp types (e.g top precision grasps, handle grasps, prismatic precision and power grasps, circular precision and power grasps, etc.) and select different grasp types based on task requirements. Specifically, as defined by Cutkosky in [1], we can define all grasp types hierarchically as sub-types of precision and power grasps. Thus just as the results presented in this letter allow us to refine a single classifier into two classifiers and improve inference performance by modeling grasp type, we expect we recursively split grasp types into additional, more refined grasp sub-types if sufficient examples of each sub-type were available. Furthermore, for planning, we could use the tree-based structure of the grasp taxonomy to search over the discrete grasp type, potentially improving the efficiency over the linear search done in this letter.

Learning more grasp types, of course, would require some way of identifying grasp sub-types at training time which could be done in a number of ways. We believe that by improving sensation on the hand to have an array of tactile sensors on the palm and fingers, we could define grasp sub-types based on where contact exists between the hand and object. These tactile-based grasp types could either be hard coded or---as we wish to examine---discovered automatically through unsupervised learning.


%



\bibliographystyle{IEEEtran}
\bibliography{grasp_ref}  

\begin{thebibliography}{10}
\providecommand{\url}[1]{#1}
\csname url@samestyle\endcsname
\providecommand{\newblock}{\relax}
\providecommand{\bibinfo}[2]{#2}
\providecommand{\BIBentrySTDinterwordspacing}{\spaceskip=0pt\relax}
\providecommand{\BIBentryALTinterwordstretchfactor}{4}
\providecommand{\BIBentryALTinterwordspacing}{\spaceskip=\fontdimen2\font plus
\BIBentryALTinterwordstretchfactor\fontdimen3\font minus
  \fontdimen4\font\relax}
\providecommand{\BIBforeignlanguage}[2]{{%
\expandafter\ifx\csname l@#1\endcsname\relax
\typeout{** WARNING: IEEEtran.bst: No hyphenation pattern has been}%
\typeout{** loaded for the language `#1'. Using the pattern for}%
\typeout{** the default language instead.}%
\else
\language=\csname l@#1\endcsname
\fi
#2}}
\providecommand{\BIBdecl}{\relax}
\BIBdecl

\bibitem{cutkosky1989grasp}
M.~R. Cutkosky, ``{On grasp choice, grasp models, and the design of hands for
  manufacturing tasks},'' \emph{{IEEE} Trans. Robot. and Autom.}, vol.~5,
  no.~3, pp. 269--279, 1989.

\bibitem{heinemann2015taxonomy}
F.~Heinemann, S.~Puhlmann, C.~Eppner, J.~{\'E}lvarez-Ruiz, M.~Maertens, and
  O.~Brock, ``{A taxonomy of human grasping behavior suitable for transfer to
  robotic hands},'' in \emph{Proc. IEEE Int. Conf. Robot. Autom.}, 2015, pp.
  4286--4291.

\bibitem{gualtieri2015using}
M.~Gualtieri, A.~Ten~Pas, K.~Saenko, and R.~Platt, ``{Using geometry to detect
  grasp poses in 3d point clouds},'' in \emph{Int. Symp. on Robot. Res.}, 2015.

\bibitem{lenz2015deep}
I.~Lenz, H.~Lee, and A.~Saxena, ``{Deep learning for detecting robotic
  grasps},'' \emph{Int. J. Robot. Res.}, vol.~34, no. 4-5, pp. 705--724, 2015.

\bibitem{gualtieri2016high}
M.~Gualtieri, A.~ten Pas, K.~Saenko, and R.~Platt, ``{High precision grasp pose
  detection in dense clutter},'' in \emph{Proc. IEEE/RSJ Int. Conf. Intell.
  Robots Syst.}, 2016, pp. 598--605.

\bibitem{pinto2016supersizing}
L.~Pinto and A.~Gupta, ``{Supersizing self-supervision: learning to grasp from
  50K tries and 700 robot hours},'' in \emph{Proc. IEEE Int. Conf. Robot.
  Autom.}, 2016, pp. 3406--3413.

\bibitem{levine2016learning}
S.~Levine, P.~Pastor, A.~Krizhevsky, J.~Ibarz, and D.~Quillen, ``{Learning
  hand-eye coordination for robotic grasping with deep learning and large-scale
  data collection},'' \emph{Int. J. Robot. Res.}, vol.~37, no. 4-5, pp.
  421--436, 2018.

\bibitem{mahler2017dex}
J.~Mahler, J.~Liang, S.~Niyaz, M.~Laskey, R.~Doan, X.~Liu, J.~A. Ojea, and
  K.~Goldberg, ``{Dex-net 2.0: deep learning to plan robust grasps with
  synthetic point clouds and analytic grasp metrics},'' in \emph{Robotics
  Science and Systems}, 2017.

\bibitem{johns2016deep}
E.~Johns, S.~Leutenegger, and A.~J. Davison, ``{Deep learning a grasp function
  for grasping under gripper pose uncertainty},'' in \emph{Proc. IEEE/RSJ Int.
  Conf. Intell. Robots Syst.}, 2016, pp. 4461--4468.

\bibitem{varley2015generating}
J.~Varley, J.~Weisz, J.~Weiss, and P.~Allen, ``{Generating multi-fingered
  robotic grasps via deep learning},'' in \emph{Proc. IEEE/RSJ Int. Conf.
  Intell. Robots Syst.}, 2015, pp. 4415--4420.

\bibitem{veres2017modeling}
M.~Veres, M.~Moussa, and G.~W. Taylor, ``{Modeling grasp motor imagery through
  deep conditional generative models},'' \emph{{IEEE} Robot. and Autom.
  Letters}, vol.~2, no.~2, pp. 757--764, 2017.

\bibitem{lu2017grasp}
Q.~Lu, K.~Chenna, B.~Sundaralingam, and T.~Hermans, ``{Planning multi-fingered
  grasps as probabilistic inference in a learned deep network},'' in \emph{Int.
  Symp. on Robot. Res.}, 2017.

\bibitem{miller2003automatic}
A.~T. Miller, S.~Knoop, H.~I. Christensen, and P.~K. Allen, ``{Automatic grasp
  planning using shape primitives},'' in \emph{Proc. IEEE Int. Conf. Robot.
  Autom.}, vol.~2, 2003, pp. 1824--1829.

\bibitem{roa2012power}
M.~A. Roa, M.~J. Argus, D.~Leidner, C.~Borst, and G.~Hirzinger, ``{Power grasp
  planning for anthropomorphic robot hands},'' in \emph{Proc. IEEE Int. Conf.
  Robot. Autom.}, 2012, pp. 563--569.

\bibitem{hang2016hierarchical}
K.~Hang, M.~Li, J.~A. Stork, Y.~Bekiroglu, F.~T. Pokorny, A.~Billard, and
  D.~Kragic, ``{Hierarchical fingertip space: A unified framework for grasp
  planning and in-hand grasp adaptation},'' \emph{{IEEE} Trans. on Robot.},
  vol.~32, no.~4, pp. 960--972, 2016.

\bibitem{zhu2004planning}
X.~Zhu and H.~Ding, ``{Planning force-closure grasps on 3-D objects},'' in
  \emph{Proc. IEEE Int. Conf. Robot. Autom.}, 2004, pp. 1258--1263.

\bibitem{feix2016grasp}
T.~Feix, J.~Romero, H.-B. Schmiedmayer, A.~M. Dollar, and D.~Kragic, ``{The
  grasp taxonomy of human grasp types},'' \emph{IEEE Trans. on Human-Machine
  Systems}, vol.~46, no.~1, pp. 66--77, 2016.

\bibitem{bohg2014data}
J.~Bohg, A.~Morales, T.~Asfour, and D.~Kragic, ``{Data-driven grasp synthesis-a
  survey},'' \emph{{IEEE} Trans. on Robot.}, vol.~30, no.~2, pp. 289--309,
  2014.

\bibitem{sahbani2012overview}
A.~Sahbani, S.~El-Khoury, and P.~Bidaud, ``{An overview of 3D object grasp
  synthesis algorithms},'' \emph{Robotics and Autonomous Systems}, vol.~60,
  no.~3, pp. 326--336, 2012.

\bibitem{Grupen1991}
R.~A. Grupen, ``{Planning grasp strategies for multifingered robot hands},'' in
  \emph{Proc. IEEE Int. Conf. Robot. Autom.}, 1991, pp. 646--651.

\bibitem{ferrari1992planning}
C.~Ferrari and J.~Canny, ``{Planning optimal grasps},'' in \emph{Proc. IEEE
  Int. Conf. Robot. Autom.}, 1992, pp. 2290--2295.

\bibitem{Dai2015}
H.~Dai, A.~Majumdar, and R.~Tedrake, ``{Synthesis and optimization of force
  closure grasps via sequential semidefinite programming},'' in \emph{Int.
  Symp. on Robot. Res.}, 2015, pp. 1--16.

\bibitem{vahrenkamp2018planning}
N.~Vahrenkamp, E.~Koch, M.~Waechter, and T.~Asfour, ``{Planning high-quality
  grasps using mean curvature object skeletons},'' \emph{{IEEE} Robot. and
  Autom. Letters}, 2018.

\bibitem{morales2006integrated}
A.~Morales, T.~Asfour, P.~Azad, S.~Knoop, and R.~Dillmann, ``{Integrated grasp
  planning and visual object localization for a humanoid robot with
  five-fingered hands},'' in \emph{Proc. IEEE/RSJ Int. Conf. Intell. Robots
  Syst.}, 2006, pp. 5663--5668.

\bibitem{Saxena-aaai2008}
A.~Saxena, L.~L.~S. Wong, and A.~Y. Ng, ``{Learning grasp strategies with
  partial shape information},'' in \emph{AAAI}, 2008, pp. 1491--1494.

\bibitem{kappler2015leveraging}
D.~Kappler, J.~Bohg, and S.~Schaal, ``{Leveraging big data for grasp
  planning},'' in \emph{Proc. IEEE Int. Conf. Robot. Autom.}, 2015, pp.
  4304--4311.

\bibitem{redmon2015real}
J.~Redmon and A.~Angelova, ``{Real-time grasp detection using convolutional
  neural networks},'' in \emph{Proc. IEEE Int. Conf. Robot. Autom.}, 2015, pp.
  1316--1322.

\bibitem{kumra2016robotic}
S.~Kumra and C.~Kanan, ``{Robotic grasp detection using deep convolutional
  neural networks},'' in \emph{Proc. IEEE/RSJ Int. Conf. Intell. Robots Syst.},
  2017.

\bibitem{osa2016experiments}
T.~Osa, J.~Peters, and G.~Neumann, ``{Experiments with hierarchical
  reinforcement learning of multiple grasping policies},'' in \emph{Int. Symp.
  on Exp. Robot.}\hskip 1em plus 0.5em minus 0.4em\relax Springer, 2016, pp.
  160--172.

\bibitem{cai2016understanding}
M.~Cai, K.~M. Kitani, and Y.~Sato, ``{Understanding hand-object manipulation
  with grasp types and object attributes},'' in \emph{Robotics Science and
  Systems}, 2016.

\bibitem{dang2014semantic}
H.~Dang and P.~K. Allen, ``{Semantic grasping: planning task-specific stable
  robotic grasps},'' \emph{Auton. Robots}, vol.~37, no.~3, pp. 301--316, 2014.

\bibitem{song2015task}
D.~Song, C.~H. Ek, K.~Huebner, and D.~Kragic, ``{Task-based robot grasp
  planning using probabilistic inference},'' \emph{{IEEE} Trans. on Robot.},
  vol.~31, no.~3, pp. 546--561, 2015.

\bibitem{Fang2018Task}
K.~Fang, Y.~Zhu, A.~Garg, V.~Mehta, A.~Kuryenkoy, L.~Fei-Fei, and S.~Savarese,
  ``{Learning task-oriented grasping for tool manipulation with simulated
  self-supervision},'' in \emph{Robotics Science and Systems}, 2018.

\bibitem{KollerPGMs2009}
D.~Koller and N.~Friedman, \emph{{Probabilistic graphical models - principles
  and techniques}}.\hskip 1em plus 0.5em minus 0.4em\relax {MIT} Press, 2009.

\bibitem{burchfiel2017eigen}
B.~Burchfiel and G.~Konidaris, ``Bayesian eigenobjects: a unified framework for
  3d robot perception,'' in \emph{Robotics Science and Systems}, 2017.

\bibitem{byrd1995limited}
R.~H. Byrd, P.~Lu, J.~Nocedal, and C.~Zhu, ``{A limited memory algorithm for
  bound constrained optimization},'' \emph{SIAM Journal on Scientific
  Computing}, vol.~16, no.~5, pp. 1190--1208, 1995.

\bibitem{zhu1997algorithm}
C.~Zhu, R.~H. Byrd, P.~Lu, and J.~Nocedal, ``{Algorithm 778: L-BFGS-B: Fortran
  subroutines for large-scale bound-constrained optimization},'' \emph{ACM
  Trans. on Mathematical Software}, vol.~23, no.~4, pp. 550--560, 1997.

\bibitem{singh2014bigbird}
A.~Singh, J.~Sha, K.~S. Narayan, T.~Achim, and P.~Abbeel, ``{Bigbird: a
  large-scale 3D database of object instances},'' in \emph{Proc. IEEE Int.
  Conf. Robot. Autom.}, 2014, pp. 509--516.

\bibitem{calli2015benchmarking}
B.~Calli, A.~Singh, A.~Walsman, S.~Srinivasa, P.~Abbeel, and A.~M. Dollar,
  ``{The YCB object and model set},'' in \emph{IEEE Int. Conf. on Advanced
  Robotics}, 2015, pp. 510--517.

\end{thebibliography}

\end{document}